\documentclass[runningheads]{llncs}
\usepackage[T1]{fontenc}
\usepackage{graphicx}
\usepackage{booktabs}
\usepackage{url}        
\usepackage{hyperref}   
\usepackage{caption}    
\begin{document}
\title{Optimizing Indoor Environmental Quality in Smart Buildings Using Deep Learning}
\titlerunning{Optimizing IEQ in Smart Buildings Using DL}

\author{Youssef Sabiri\thanks{Corresponding author.}\inst{1} \and
Walid Houmaidi\inst{1} \and
Aaya Bougrine\inst{1} \and
Salmane El Mansour Billah\inst{1}}
\authorrunning{Y. Sabiri et al.}
\institute{Al Akhawayn University, Ifrane, Morocco\\
\email{Y.sabiri@aui.ma}
}
\maketitle              
\begin{abstract}
Ensuring optimal Indoor Environmental Quality (IEQ) is vital for occupant health and productivity, yet it often comes at a high energy cost in conventional Heating, Ventilation, and Air Conditioning (HVAC) systems. This paper proposes a deep learning–driven approach to proactively manage IEQ parameters—specifically CO\textsubscript{2} concentration, temperature, and humidity—while balancing building energy efficiency. Leveraging the ROBOD dataset collected from a net-zero energy academic building, we benchmark three architectures—Long Short-Term Memory (LSTM), Gated Recurrent Units (GRU), and a hybrid Convolutional Neural Network–LSTM (CNN-LSTM)—to forecast IEQ variables across various time horizons. Our results show that GRU achieves the best short-term prediction accuracy with lower computational overhead, whereas CNN-LSTM excels in extracting dominant features for extended forecasting windows. Meanwhile, LSTM offers robust long-range temporal modeling. The comparative analysis highlights that prediction reliability depends on data resolution, sensor placement, and fluctuating occupancy conditions. These findings provide actionable insights for intelligent Building Management Systems (BMS) to implement predictive HVAC control, thereby reducing energy consumption and enhancing occupant comfort in real-world building operations.

\end{abstract}

\section{Introduction}

IEQ is a crucial factor affecting occupant health, comfort, and productivity. According to the U.S. Environmental Protection Agency (EPA), individuals spend approximately 90\% of their time indoors, where pollutant concentrations can be 2 to 5 times higher than outdoor levels, increasing the risk of respiratory diseases, heart disease, and cancer~\cite{EPA}. Similarly, the World Health Organization (WHO) reported that household air pollution contributed to 3.2 million premature deaths globally in 2020, including 237{,}000 deaths among children under five, emphasizing the urgent need for improved indoor air management~\cite{WHO}. Key factors influencing IEQ include CO\textsubscript{2} levels, volatile organic compounds (VOCs), humidity, and temperature, all of which impact cognitive function and workplace efficiency. Poor IEQ has been associated with sick building syndrome, asthma, and neurobehavioral disorders. HVAC systems play a significant role in maintaining IEQ but account for 40\% of a building’s total energy consumption, necessitating more efficient operational strategies~\cite{Australia,ref1,ref2}. Traditional HVAC systems operate on fixed schedules and reactive adjustments, often resulting in delayed responses and increased energy consumption. To address these limitations, predictive modeling has emerged as a promising alternative, leveraging real-time data analytics to enhance HVAC efficiency while maintaining optimal IEQ~\cite{ref3}.

This study investigates the effectiveness of deep learning models—LSTM, GRU, and CNN—for forecasting key IEQ parameters. Specifically, it focuses on developing and validating deep learning approaches for real-time CO\textsubscript{2}, humidity, and temperature forecasting using the ROBOD dataset~\cite{dataset}; benchmarking model accuracy, computational efficiency, and robustness across different forecasting horizons; assessing the impact of sensor data granularity and feature selection on predictive performance; and providing insights into integrating predictive IEQ models into intelligent HVAC management systems.

By systematically evaluating these models, this study contributes to the advancement of smart building technologies, enabling sustainable and adaptive climate control strategies that enhance energy efficiency and occupant well-being.
\section{Literature Review}
Recent advancements in IEQ forecasting have increasingly utilized machine learning (ML) and deep learning techniques to enhance predictive accuracy. Traditional models such as Multiple Linear Regression (MLR) and Autoregressive Integrated Moving Average (ARIMA) have been used to predict indoor air pollutant concentrations and temperature trends, yet they often fail to capture the complex nonlinear dependencies inherent to indoor environments. More advanced machine learning methods like Random Forests, Gradient Boosting, and Support Vector Machines have been applied for predicting CO\textsubscript{2} levels and optimizing HVAC performance, although their effectiveness is still limited by issues such as feature selection and sensor placement~\cite{ref4}.

Deep learning models—including LSTM networks, GRU, and CNN—have demonstrated superior capabilities in capturing long-term dependencies and complex patterns in IEQ data. For example, hybrid models that combine CNN and LSTM layers have effectively predicted particulate matter concentrations in subway stations, where both spatial and temporal dynamics are crucial~\cite{ref5}. However, challenges remain in optimizing sensor placement and integrating real-time predictive models into BMS for proactive HVAC control~\cite{ref6}. While emerging approaches such as federated and transfer learning offer promising directions, comprehensive comparative analyses of deep learning architectures in real-world BMS deployments are still limited~\cite{ref6,ref8}.

Recent work has extended deep learning applications to HVAC system optimization and indoor microclimate forecasting. For instance, the paper~\cite{new1} proposed an energy consumption optimization method for HVAC systems based on deep reinforcement learning; their experiments showed that a CNN–LSTM model for HVAC energy consumption prediction outperformed baseline models while reducing training time. Similarly, the authors of~\cite{new2} developed an H-Ahead multivariate microclimate forecasting system using a GRU model to accurately predict indoor temperature, humidity, and CO\textsubscript{2} levels. In addition, the paper~\cite{new3} introduced a digital twin framework for indoor temperature prediction in smart buildings by leveraging LSTM and BiLSTM models. For residential HVAC systems, authors of~\cite{new4} applied deep learning to optimize energy consumption while maintaining occupant comfort, and~\cite{new5} integrated multi-objective optimization into a deep learning algorithm for forecasting energy demands in smart homes. The applicability of deep learning for indoor temperature prediction was further demonstrated by~\cite{new6} in an educational building case study. On the control side, deep reinforcement learning was employed by~\cite{new7} to achieve energy-efficient thermal comfort control in smart buildings, while~\cite{new9} optimized multi-regional commercial HVAC systems using DRL. Moreover, a supervised-learning strategy for optimal demand response in multi-zone office HVAC systems was proposed in~\cite{new13}, and finally, an entropy-driven deep reinforcement learning method for HVAC system optimization was introduced in~\cite{new19}. Together, these studies underscore the growing impact of deep learning methods in both predictive IEQ modeling and advanced HVAC control.

\section{Methods}

\subsection{Proposed System Architecture}
In the envisioned smart building scenario, each room is equipped with an indoor air quality sensor capturing temperature, CO\textsubscript{2}, and humidity data in real time. These measurements stream to a central or edge-based deep learning model that predicts IEQ values for the subsequent five-minute interval. The HVAC system then proactively adjusts temperature setpoints, ventilation rates, and humidity levels to maintain an optimal indoor environment. When low latency or on-premise processing is required, sensor data can be processed locally on an edge device; otherwise, it can be uploaded to a cloud server for more extensive computation. This closed-loop mechanism enhances both occupant comfort and energy efficiency.

\subsection{Research Question}
While the end goal of this study is to facilitate a robust, smart-building ecosystem capable of real-time, proactive HVAC control, the core focus of our work lies in developing and benchmarking deep learning models for IEQ forecasting. In particular, we seek to identify which model architecture—given extensive training experiments, hyperparameter tuning, and diverse design considerations—can most accurately predict future CO\textsubscript{2}, temperature, and humidity values.

\subsection{Dataset Characteristics and Preprocessing}
\noindent \textbf{The ROBOD (Room-level Occupancy and Building Operation Dataset)} used in this study was collected from five distinct rooms within a university environment, including two lecture rooms, an administrative office, a research office, and a library. These rooms capture diverse occupancy patterns and building operations. The dataset comprises 123{,}789 valid records. It integrates indoor environmental conditions, energy consumption, HVAC operations, outdoor weather conditions, Wi-Fi device counts, and ground-truth occupancy information~\cite{dataset}.

\subsubsection{Data Cleaning and Preparation}
In the preprocessing stage, missing values were handled using polynomial interpolation (order=3) to preserve the time-series integrity. This method was chosen over simpler techniques, such as linear interpolation, for its ability to better reflect fluctuations and trends, ensuring a more accurate estimation of missing values.

\subsubsection{Feature Selection}
For this study, three IEQ features were chosen to keep the model interpretable while focusing on occupant well-being and energy usage:
\begin{itemize}
    \item \textbf{Air Temperature (°C):} Influences occupant comfort and HVAC energy consumption.
    \item \textbf{Indoor CO\textsubscript{2} Levels (ppm):} Serves as an indicator of indoor air quality and occupancy.
    \item \textbf{Humidity (\%RH):} Affects occupant comfort and the likelihood of mold or dryness issues.
\end{itemize}
Their inclusion allows for actionable insights into both occupant well-being and energy optimization~\cite{ref1}.

\subsubsection{Addressing Time Gaps and Extracting Continuous Intervals}
The dataset contained significant time gaps, sometimes spanning weeks or months, due to sensor malfunctions and irregular logging. To address this, only continuous, uninterrupted intervals were retained after standardizing timestamps and discarding inconsistent entries to maintain the sequential nature of the data.

\begin{figure}[!htb]
\centering
\includegraphics[width=\textwidth]{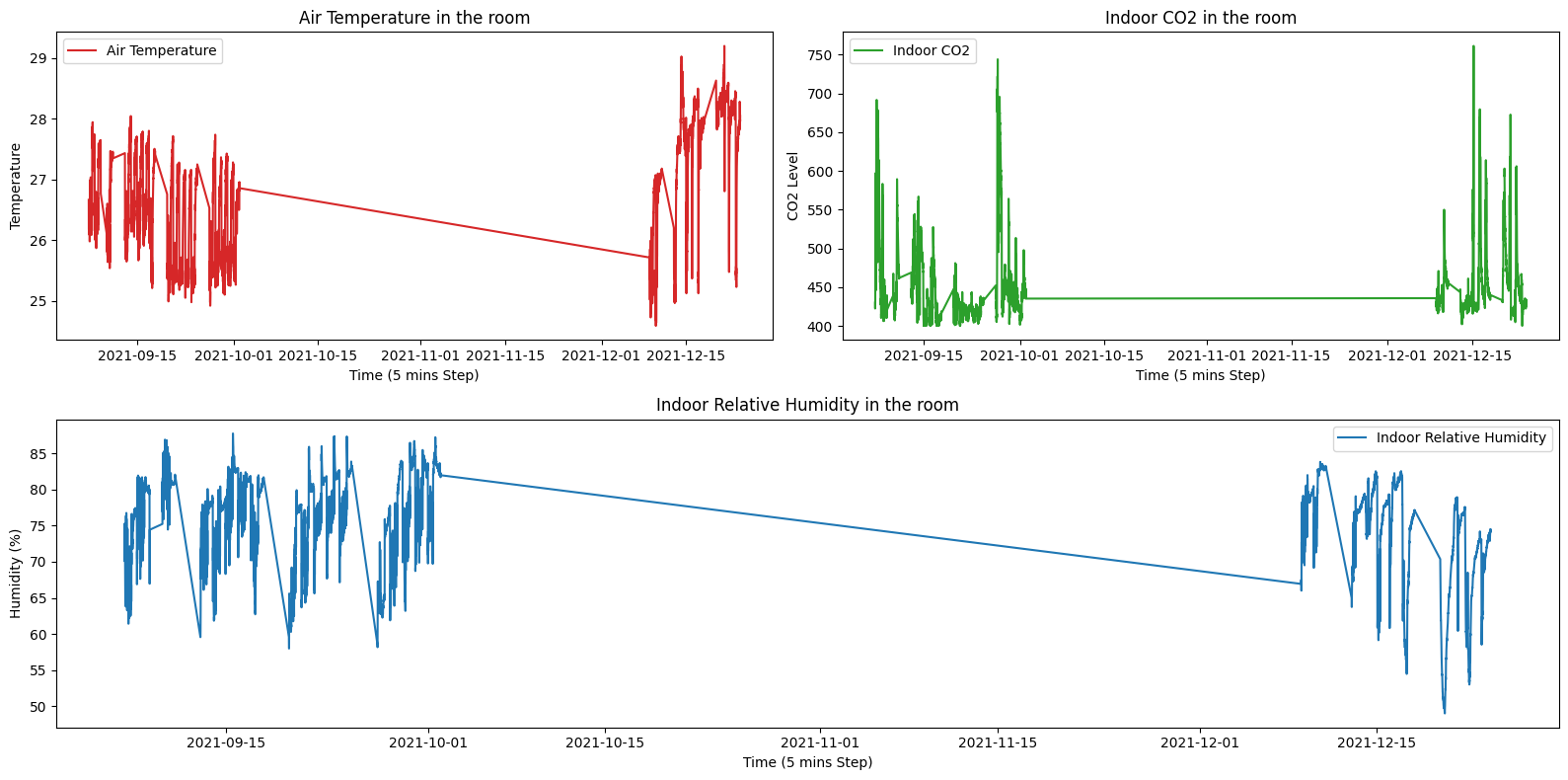}
\caption{Visual inspection of temperature, CO\textsubscript{2}, and humidity time series revealing substantial intervals with missing data (see flat lines and sudden jumps).}
\label{fig:time_gaps}
\end{figure}

Figure~\ref{fig:time_gaps} shows flat regions and abrupt transitions caused by sensor downtime and logging issues. Removing these sections allowed the model to capture meaningful temporal dependencies.

\begin{figure}[!htb]
\centering
\includegraphics[width=1\textwidth]{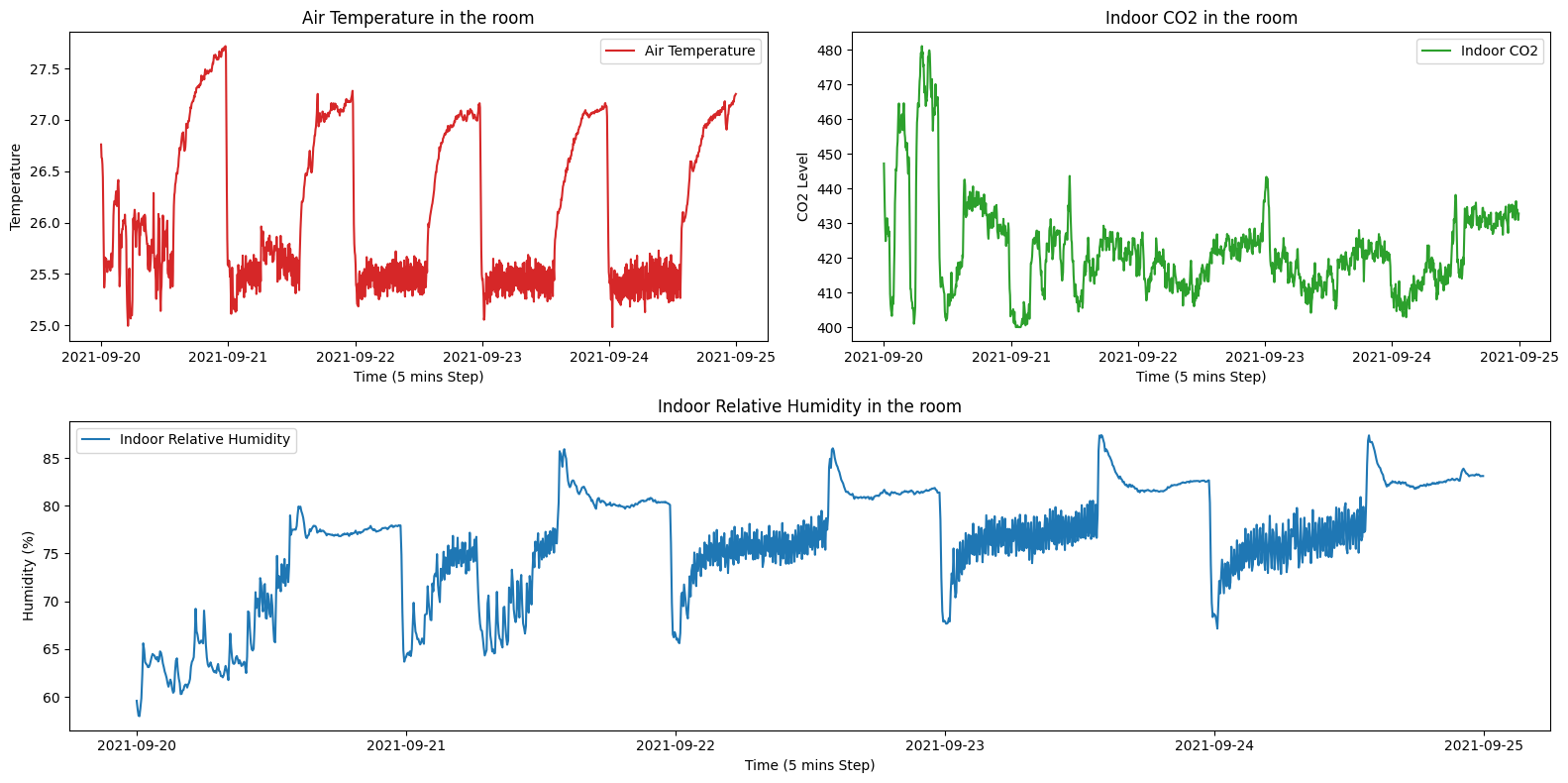}
\caption{Example of a good continuous interval showing clear daily patterns in air temperature, indoor CO\textsubscript{2}, and humidity.}
\label{fig:cont_good}
\end{figure}

\begin{figure}[!htb]
\centering 
\includegraphics[width=\textwidth]{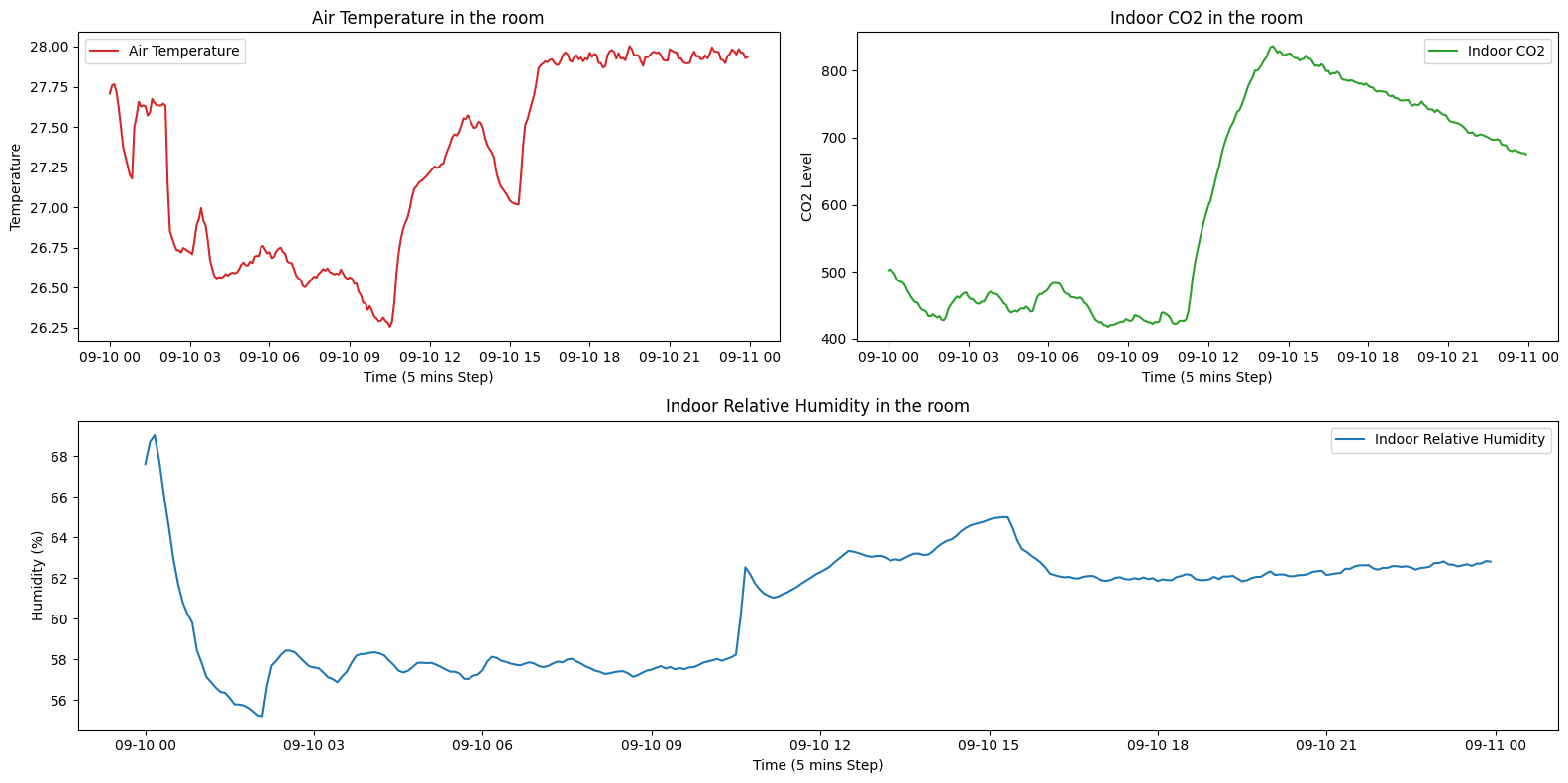}
\caption{Example of a continuous interval with irregular or noisy readings, offering fewer obvious patterns.}
\label{fig:cont_noisy}
\end{figure}

\paragraph{Examples of Extracted Continuous Intervals}
After eliminating gaps, distinct segments were identified:
\begin{itemize}
\item \textbf{Segments with clear patterns:} Figure~\ref{fig:cont_good} shows intervals with observable trends, such as daily temperature cycles or CO\textsubscript{2} spikes.
\item \textbf{Noisy segments:} Some intervals (Figure~\ref{fig:cont_noisy}) display irregular readings with unclear trends, requiring the model to manage stochastic variations.
\end{itemize}

\subsubsection{(4) Feature Transformation}
To enhance the dataset's predictive power for time series modeling, several feature transformations were applied. The timestamp was decomposed into cyclical features—\textit{day\_sin, day\_cos, month\_sin, month\_cos}—to capture recurring daily and monthly trends in temperature, CO2 levels, and humidity. This transformation helped neural networks effectively interpret temporal patterns.

Normalization was performed by scaling all features to a [0,1] range, ensuring consistent feature scaling and improved model convergence. Additionally, a sliding window approach with a window size of 12 (1 hour of 5-minute intervals) was used to capture short-term dependencies, providing the model with past observations for better future predictions.

\subsubsection{(5) Dataset Splits}
Finally, the dataset was split into training (85\%), validation (7.5\%), and test (7.5\%) sets. This ratio provides the model with sufficient training data while retaining adequate unseen data for validation and testing, minimizing the risk of overfitting and ensuring robust performance evaluations.
\subsection{Model Selection and Justifications}
In this study, three deep learning architectures---LSTM, GRU, and a hybrid CNN-LSTM---were selected to forecast indoor CO\textsubscript{2} levels, temperature, and relative humidity over varying time horizons (daily, weekly, and monthly). Each model addresses different aspects of temporal dependency, computational complexity, and feature extraction within time-series data.

\subsubsection{LSTM Model}
LSTM networks leverage gating mechanisms (input, forget, output) to capture long-range dependencies in sequential data. By retaining relevant contextual information over extended sequences, they are well-suited for IEQ forecasting, where parameters exhibit seasonal, daily, and occupancy-driven fluctuations.
\subsubsection{GRU Model}
GRUs function similarly to LSTMs but with fewer internal gates (update, reset), often resulting in faster training and lower memory usage. They deliver performance comparable to LSTMs and are especially advantageous where computational resources or latency constraints demand frequent model updates.
\subsubsection{Hybrid CNN-LSTM Model}
A combined CNN-LSTM architecture unites convolutional layers (for extracting local temporal patterns) with recurrent layers (for modeling longer-term dependencies). This approach is valuable for multi-scale IEQ forecasting tasks in net-zero energy buildings, where both short-lived anomalies and seasonal shifts influence comfort and efficiency.
\subsubsection{Comparative Rationale and Expected Outcomes}
Using LSTM, GRU, and hybrid CNN-LSTM allows a thorough exploration of trade-offs among predictive accuracy, computational efficiency, and model complexity for IEQ forecasting.
\subsection{Training Details and Hyperparameters}
All models were trained under identical conditions to ensure a fair comparison:
\begin{itemize}
    \item \textbf{Window Size (12)}: Each model receives 12 consecutive time steps (one hour of data) for forecasts.
    \item \textbf{Batch Size (64)}: Ensures balanced memory usage and stable gradient updates.
    \item \textbf{Epochs (100)}: Training stops early if validation loss fails to improve for 7 epochs, otherwise it continues to a max of 100 epochs.
    \item \textbf{Optimizer (Adam)}: Chosen for robust performance and adaptability to sparse gradients.
    \item \textbf{Learning Rate (0.0001)}: Starting at 0.0001 then reduced by half (down to $10^{-6}$) whenever validation loss stagnates for 3 consecutive epochs.
    \item \textbf{Loss Function \& Metrics}: Mean Absolute Error (MAE) as the primary loss function; Root Mean Squared Error (RMSE) tracked to highlight larger deviations.
\end{itemize}

\section{Results and Discussion}
\subsection{Performance Comparison}
After conducting training experiments detailed above, all models were evaluated on the test set. The table presents a comparative analysis of performance metrics—MAE, MSE, RMSE, and R\textsuperscript{2}—for three models (\textbf{LSTM}, \textbf{GRU}, and \textbf{Hybrid CNN-LSTM}) across global metrics and specific parameters: air temperature, indoor CO$_{2}$ levels, and indoor relative humidity. Metrics used for evaluation are MAE, Mean Squared Error (MSE), RMSE, and Coefficient of Determination ($R^2$).

\begin{table}[ht]
\centering
\caption{Performance Metrics of LSTM, GRU, and Hybrid CNN-LSTM Models}
\label{tab:performance_metrics}
\begin{tabular}{lccc ccc ccc ccc}
\hline
\textbf{Metric} & \multicolumn{3}{c}{\textbf{Global Metrics}} & \multicolumn{3}{c}{\textbf{Air Temperature}} & \multicolumn{3}{c}{\textbf{Indoor CO\textsubscript{2}}} & \multicolumn{3}{c}{\textbf{Humidity}} \\
\hline
& \textbf{LSTM} & \textbf{GRU} & \textbf{Hybrid} & \textbf{LSTM} & \textbf{GRU} & \textbf{Hybrid} & \textbf{LSTM} & \textbf{GRU} & \textbf{Hybrid} & \textbf{LSTM} & \textbf{GRU} & \textbf{Hybrid} \\
\hline
MAE  & 1.3332 & 1.0320 & 9.1651 & 0.0644 & 0.0362 & 0.1470 & 3.6760 & 2.8950 & 26.6146 & 0.2591 & 0.1646 & 0.7337 \\
MSE  & 7.8433 & 4.5946 & 314.5121 & 0.0102 & 0.0032 & 0.0368 & 23.3138 & 13.6871 & 942.5850 & 0.2059 & 0.0935 & 0.9144 \\
RMSE & 2.8006 & 2.1435 & 17.7345 & 0.1010 & 0.0566 & 0.1919 & 4.8284 & 3.6996 & 30.7015 & 0.4538 & 0.3058 & 0.9562 \\
R\textsuperscript{2} & 0.9804 & 0.9915 & 0.7078 & 0.9747 & 0.9921 & 0.9087 & 0.9823 & 0.9896 & 0.2849 & 0.9842 & 0.9928 & 0.9298 \\
\hline
\end{tabular}
\end{table}
\begin{figure}[h]
\centering
\includegraphics[width=1.05\textwidth]{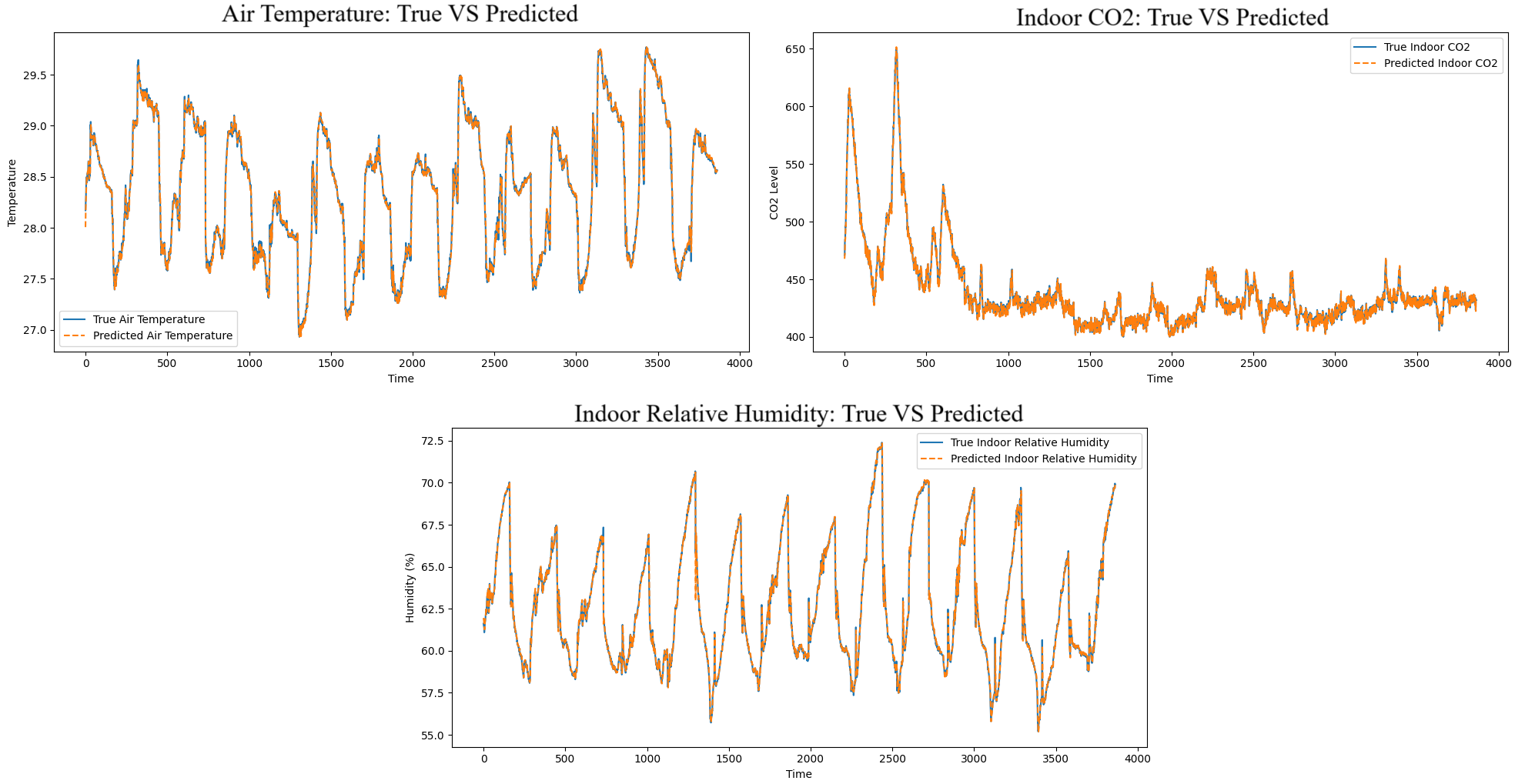}
\caption{GRU model predictions (dashed orange) vs. ground-truth measurements (solid blue) for air temperature, CO\textsubscript{2}, and humidity on the test set.}
\label{fig:gru_predictions}
\end{figure}
\subsection{Interpretation of Results}
Evaluation of LSTM, GRU and CNN-LSTM models reveals distinct differences in predicting IEQ parameters, specifically in terms of MAE and $R^2$. The GRU model outperforms the others, achieving the lowest Global MAE of 1.0320 and a high $R^2$ value of 0.9915, indicating superior predictive accuracy. For air temperature, the GRU model attains an MAE of 0.0362 with an $R^2$ of 0.9921. It also performs well in predicting indoor CO\textsubscript{2} levels (MAE: 2.8950, $R^2$: 0.9896) and relative humidity (MAE: 0.1646, $R^2$: 0.9928).

The LSTM model demonstrates competitive performance, with a Global MAE of 1.3332 and an $R^2$ of 0.9804. It achieves an MAE of 0.0644 ($R^2$: 0.9747) for air temperature, 3.6760 ($R^2$: 0.9823) for CO\textsubscript{2}, and 0.2591 ($R^2$: 0.9842) for humidity. 

In contrast, the CNN-LSTM model exhibits higher error rates, with a Global MAE of 9.1651 and a lower $R^2$ of 0.7078. It particularly struggles with CO\textsubscript{2} prediction, showing an MAE of 26.6146 and an $R^2$ of 0.2849, indicating challenges in capturing the underlying data patterns.

In general, the GRU model offers the best balance of accuracy and efficiency, making it the preferred choice for forecasting IEQ parameters. The comparatively inferior performance of the CNN-LSTM model can be attributed to its complexity, which could lead to overfitting or difficulties in modeling temporal dependencies effectively. Figure~\ref{fig:gru_predictions} shows GRU's superior performance on the test set.

\section{Conclusion and Future Work}
In this study, we investigated the effectiveness of deep learning models, specifically LSTM, GRU, and CNN architectures, to forecast IEQ parameters using the ROBOD dataset. Our results indicate that GRU provides an optimal balance between prediction accuracy and computational efficiency for short-term forecasting, while CNN demonstrates superior performance in feature extraction for long-term predictions. LSTM, with its ability to capture long-range dependencies, performed well in scenarios with complex temporal patterns. These findings contribute to the ongoing development of intelligent BMS that aims to enhance occupant comfort while optimizing energy consumption.

These promising results could benefit from additional future work. First, the integration of additional environmental and contextual features, such as weather forecasts and occupancy patterns, could further enhance model accuracy. Second, exploring advanced architectures such as transformers and hybrid models may provide deeper insights into complex IEQ dynamics. Furthermore, our study did not directly explore real-time deployment of these models in smart building environments; however, our findings could be further confirmed by the evaluation of their robustness under dynamic real-time user conditions. Finally, developing lightweight models suitable for edge computing environments can facilitate live adaptive HVAC control, ultimately contributing to more sustainable and energy-efficient building operations.

It is important to note that practical implementation faces limitations including the computational constraints of edge devices, challenges in sensor placement and data quality, and the variability of occupancy patterns, all of which require careful consideration in future research.
\begin{credits}
\subsubsection{\discintname}
The authors have no competing interests to disclose.
\end{credits}
\bibliographystyle{splncs04_unsrt}
\bibliography{references}         

\end{document}